\documentclass[11pt]{article}

\usepackage[preprint]{acl}

\usepackage{times}
\usepackage{latexsym}

\usepackage[T1]{fontenc}
\usepackage[utf8]{inputenc}

\usepackage{microtype}

\usepackage{inconsolata}

\usepackage{graphicx}
\usepackage{booktabs} 
\usepackage{float}

\usepackage{multirow}
\usepackage{booktabs} % Required for \toprule, \midrule, and \bottomrule
\usepackage{amsmath}
\usepackage{subcaption} % Required for subfigure environment

\title{Knowledge Knows, Verbalization Tells: Disentangling Latent Directions for Mathematical Solvability in LLMs}

\author{
\begin{tabular}{ccc}
Nikolaos Xiros\thanks{Equal contribution. Author order determined by coin flip.} &
Maria-Eleni Zoumpoulidi\footnotemark[1] &
Georgios Paraskevopoulos
\end{tabular}
\\[0.6em]
{\normalfont Institute for Language and Speech Processing, Athena Research Center, Greece}\\
{\normalfont \texttt{\{n.xiros,m.zoumpoulidi,g.paraskevopoulos\}@athenarc.gr}}
}

\begin{document}

\maketitle
\begin{abstract}
Although LLMs have made significant progress in mathematical reasoning, determining whether a mathematical problem is solvable remains a fundamental yet challenging capability. While recent studies have probed internal representations of model solvability beliefs, verbalization has primarily been studied behaviorally rather than as an internal representation, limiting its analysis and manipulation. We address this gap by separately probing representations of solvability knowledge and verbalization, allowing us to disentangle the two within model hidden states. Across multiple LLMs, we show that knowledge and verbalization are encoded as distinct, linearly decodable representations and that fabrication is primarily associated with changes in verbalization rather than the underlying knowledge. Prompting with unsolvability cues reduces fabrication primarily by shifting verbalization, while activation steering demonstrates that these representations can be mechanistically manipulated to improve model abstention.
\end{abstract}

%------------------------------------------------------------------
\section{Introduction}
%------------------------------------------------------------------

\begin{figure}[t]
\centering
\includegraphics[width=1.0\columnwidth]{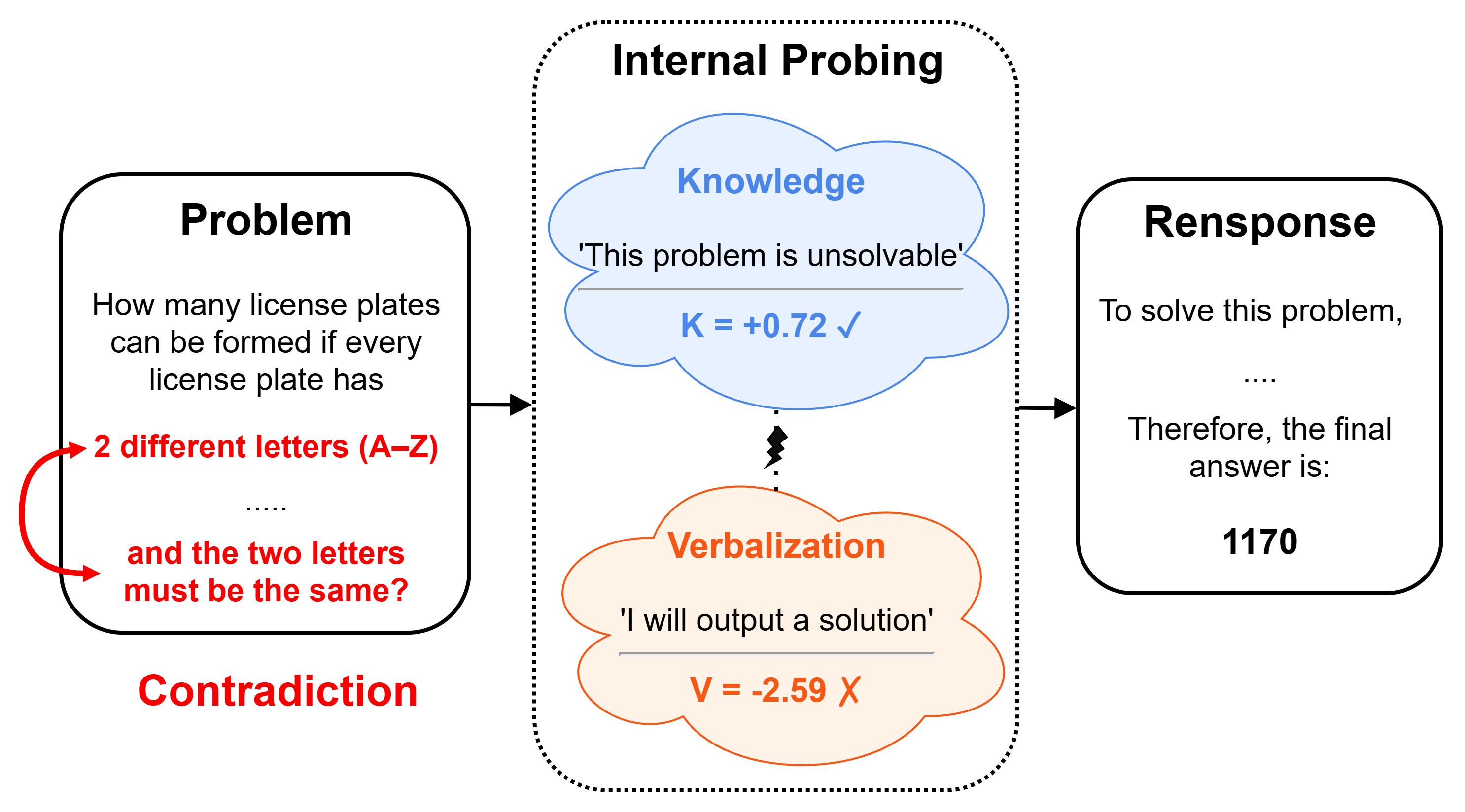}
\caption{Overview of our approach. Given a problem, we probe two complementary directions in LLM representations: knowledge, capturing whether the model internally recognizes the problem as unsolvable, and verbalization, capturing whether it explicitly communicates this judgment. The final generated response typically follows the verbalization direction.}
\label{fig:overview}
\end{figure}

Assessing the solvability of a problem is a core yet challenging aspect of mathematical reasoning, even for experts, motivating a growing body of research on large language models' (LLMs) ability to recognize unsolvable problems. 
\\
Existing work spans the development of benchmarks and evaluation of prompting techniques \citep{xue2025reliablemathbenchmarkreliablemathematical}, as well as mechanistic interpretability approaches that probe the internal and geometric representations of solvability beliefs \citep{liu2026answeringunanswerableerrknowingly,sanyal2025confidencecompetence}. Such analyses suggest that models often possess latent knowledge about a problem's unsolvability but fail to express it in their responses, while also identifying interventions that can mitigate this behavior \citep{liu2026answeringunanswerableerrknowingly,sanyal2025confidencecompetence}.
\\
%In parallel, research on answerability in broader, primarily non-mathematical domains has evolved along similar methodological lines, encompassing benchmarks \citep{kirichenko2026abstentionbench}, probing-based analyses \citep{slobodkin2023curiouscasehallucinatoryunanswerability}, and causal approaches that identify abstention-related mechanisms through activation interventions \citep{lavi-etal-2026-detecting}. These parallels suggest that insights from one setting may naturally transfer to the other.
In parallel, research on answerability in broader, primarily non-mathematical domains has evolved along similar methodological lines (\citet{kirichenko2026abstentionbench}, \citet{slobodkin2023curiouscasehallucinatoryunanswerability}, \citet{lavi-etal-2026-detecting}). These parallels suggest that insights from one setting may naturally transfer to the other.
\\
%More broadly, the question of whether a reasoning model’s verbalized chain of thought (CoT) faithfully reflects the computations that drive its answers extends beyond solvability and answerability. Research in this area ranges from benchmarking faithfulness \citep{shen2026faithcotbenchbenchmarkinginstancelevelfaithfulness} to demonstrating that chains of thought often diverge from the underlying computations that produce an answer, instead functioning as post-hoc rationalizations \citep{turpin2023language,lanham2023measuringfaithfulnesschainofthoughtreasoning}. Complementing these results, works that models frequently fail to acknowledge information or hints that influenced their reasoning \citep{chen2025reasoningmodelsdontsay,mirtaheri2026catchingrationalizationactdetecting}. Mechanistic studies using probing further suggest that models can internally represent the correct answer while ultimately producing a different prediction, identifying distinct representations of knowledge and prediction whose misalignment can lead to incorrect outputs \citep{park2026bridgingknowledgepredictiongapllms}.
More broadly, prior work on chain-of-thought faithfulness shows that models' verbalized reasoning often diverges from their underlying computations \citep{turpin2023language,lanham2023measuringfaithfulnesschainofthoughtreasoning}, fails to acknowledge influential information \citep{chen2025reasoningmodelsdontsay,mirtaheri2026catchingrationalizationactdetecting}, and can arise from a misalignment between internal knowledge and prediction representations \citep{park2026bridgingknowledgepredictiongapllms,shen2026faithcotbenchbenchmarkinginstancelevelfaithfulness}.
\\
%Although these works collectively advance our understanding of solvability detection in LLMs and reveal a misalignment between internal knowledge and verbalized reasoning, they study verbalization exclusively through textual reasoning traces. As a result, they provide limited insight into how verbalization emerges within a model's internal representations, how it interacts with the model's underlying knowledge, and how these representations might be manipulated to improve solvability detection. To address this gap, we disentangle representations of solvability knowledge from representations of verbalization, enabling us to study their interaction and causal influence on model behavior.
%thereby leaving unexplored the internal interplay between solvability knowledge and its verbalization, and limiting our ability to localize where and how this misalignment emerges within the model. 
Although these works advance our understanding of solvability detection and reveal a gap between internal knowledge and verbalized reasoning, they study verbalization only through textual reasoning traces,  hindering its analysis and manipulation. We instead disentangle internal representations of solvability knowledge and verbalization, enabling us to study their interaction and  influence on model behavior.
%We instead disentangle internal representations of solvability knowledge and verbalization, enabling us to study their interaction and  influence on model behavior.
Inspired by \citet{park2026bridgingknowledgepredictiongapllms}, who apply this framework to general multiple-choice QA scenarios, we study the interplay between these two representations in LLMs. Specifically, we probe two complementary directions: knowledge, capturing whether the model internally recognizes that a problem is unsolvable, and verbalization, capturing whether it explicitly communicates this judgment. An overview of our framework can be found in Figure ~\ref{fig:overview}. We aim to answer the following research questions: 
\textbf{RQ1.} Are solvability knowledge and solvability verbalization represented as distinct directions in model hidden states? \textbf{RQ2.} If so, how are these directions related with fabrication? \textbf{RQ3.} Can prompting influence model fabrication and the underlying representations of solvability knowledge and verbalization? \textbf{RQ4.} How do representations of solvability knowledge and solvability verbalization differ between internal reasoning traces and final output traces in Large Reasoning Models (LRMs)? \textbf{RQ5.} Can solvability knowledge and verbalized solvability judgments be mechanistically steered to improve solvability detection?

%\textbf{RQ1.} To what extent do language models internally recognize that a mathematical problem is unsolvable without verbalizing this knowledge, and can prompt interventions reduce this gap? \textbf{RQ2.} For reasoning models, how does the relationship between solvability knowledge and its verbalization differ between internal thinking traces and final output traces? \textbf{RQ3.} How do solvability knowledge and its verbalization evolve throughout the reasoning process, and how does their alignment change across reasoning depth? \textbf{RQ4.} Do prompt interventions move verbalized judgments closer to a model's latent solvability beliefs, or do they merely increase compliance with the provided hint? \textbf{RQ5.} Can solvability knowledge, solvability verbalization, and the alignment between them be selectively manipulated through representational steering interventions? 

Our main contributions are:
\begin{itemize}
    \item We introduce representation-level analysis of \textit{verbalization} and disentangle it from internal representations of solvability \textit{knowledge} in LLM hidden states, showing that the two form distinct, linearly decodable directions.
    \item We provide a comprehensive analysis of the relationship between these representations, showing that fabrication is associated with their misalignment, primarily reflects changes in verbalization rather than knowledge, and that prompting mainly shifts verbalization while reasoning and output traces exhibit distinct dynamics.
    \item We show that these directions can be selectively steered to improve solvability detection.    
\end{itemize}

%------------------------------------------------------------------
\section{Related Work}
\label{sec:related_work}
%------------------------------------------------------------------
\subsection{LLMs and Solvability-Answerability}
A growing body of research studies LLM behavior on unanswerable and unsolvable questions. \citet{xue2025reliablemathbenchmarkreliablemathematical} and \citet{kirichenko2026abstentionbench} introduce benchmarks in mathematics and across multiple domains, respectively, and show that models frequently fail to recognize unsolvability and unanswerability and instead fabricate solutions, a tendency that can be mitigated through appropriate prompting. \citet{kirichenko2026abstentionbench} further find that reasoning fine-tuning degrades abstention. \citet{liu2026answeringunanswerableerrknowingly} investigate LRMs' behavior on unsolvable math problems and probe internal activations, showing that a signal of answerability emerges and strengthens throughout the reasoning trace. Their results suggest that models often represent that a problem is unsolvable yet still fail to abstain, a behavior associated with lower confidence in abstention. To address this issue, they propose a cognitive monitoring and inference-time intervention method that improves the model's ability to abstain from answering unsolvable questions. \citet{sanyal2025confidencecompetence} identify a linearly decodable solvability-belief direction in LLM representations via probing, uncover a geometric separation between high-dimensional assessment representations and lower-dimensional execution dynamics, and show that steering activations along this axis leaves final answers unchanged, providing evidence for a decoupled "assessor" and "executor". Moving beyond mathematics, \citet{slobodkin2023curiouscasehallucinatoryunanswerability} provided a foundational result for extractive QA: linear probing separates answerable from unanswerable questions, the resulting subspace generalizes across QA datasets, and erasure of this subspace degrades abstention. Crucially, they also showed that adding a prompt hint mentioning the possibility of unanswerability improves abstention F1 by up to 80 points and visibly reorganizes the hidden-state geometry. \citet{lavi-etal-2026-detecting} identify answerability directions using a criterion based on the effect of activation additions on abstention behavior, rather than probe accuracy, and demonstrate direct control over abstention through activation addition and ablation. 

\subsection{CoT Faithfulness}
Whether a reasoning model's verbalized chain of thought reflects the computation that drives its answer is a now-established concern. \citet{turpin2023language} and \citet{lanham2023measuringfaithfulnesschainofthoughtreasoning} showed through behavioral and causal interventions that CoT explanations often act as post-hoc rationalizations rather than the actual reasoning process. \citet{chen2025reasoningmodelsdontsay} extended this to thinking models via a hint-injection paradigm and found that reasoning models rarely verbalize their use of influential hints. \citet{mirtaheri2026catchingrationalizationactdetecting} find that probes applied before any CoT tokens are generated are as effective at predicting the use of injected hints as an LLM-based CoT monitor that observes the full reasoning trace while probes applied after CoT generation exceed the performance of the same monitor. At the benchmark level, FaithCoT-Bench \citep{shen2026faithcotbenchbenchmarkinginstancelevelfaithfulness} formalizes instance-level CoT faithfulness evaluation across general reasoning tasks excluding solvability. \citet{park2026bridgingknowledgepredictiongapllms} train two linear probes: a knowledge probe predicting the ground-truth answer to a multiple-choice QA question, and a prediction probe predicting the model's actual choice. Their findings indicate that incorrect predictions arise from a misalignment between these two bases in the residual stream. Their KAPPA intervention realigns the prediction coordinate to the knowledge coordinate at inference, closing the gap. Our work is more closely related to this line of work than to the classical notion of faithfulness, which concerns whether a model's verbalized reasoning faithfully reflects its underlying computation: we study the alignment between knowledge and verbal representations, asking whether a model's internal knowledge is faithfully reflected in its internal verbal representations.

%------------------------------------------------------------------
\section{Methodology}
\label{sec:methodology}
%------------------------------------------------------------------
Our pipeline consists of the following stages: CoT generation and hidden-state extraction, data preparation for probing, probing, analysis, and steering; we describe each stage in detail below.
\paragraph{CoT generation and hidden states extraction:} We prompt the models and store the text CoT and the correspoding hidden states. The prompt used can be found in Appendix ~\ref{sec:prompts}.

\paragraph{Data Preparation for Probing:} 

Each hidden-state sequence is represented by 20 uniformly sampled token vectors, and probing is performed at the token level. More details can be found in \ref{sec:appendix_data_prep}. Our probes require two labels for each trace: (1) the ground-truth solvability of the corresponding problem, which serves as a proxy for the model's underlying knowledge, and (2) the model's CoT verbalized verdict. While the former is directly available from the dataset, determining the latter is more challenging, as unsolvable problems often induce complex reasoning trajectories in which the model may revise its assessment of solvability during the trace. We therefore adopt an LLM-as-a-judge approach, using Llama-3.3-70B-Instruct \citep{grattafiori2024llama3herdmodels} to identify the dominant behavior exhibited throughout the trace, namely, whether the model primarily attempts to solve the problem or concludes that it is unsolvable. The prompt used can be found in Appendix ~\ref{sec:prompts}. The judge was validated through human annotation on a subset of 100 samples, achieving a high Cohen's $\kappa$ (see Appendix~\ref{sec:appendix_human_annotation}).

\paragraph{Probing:} 

We train linear probes on model hidden states to assess the extent to which solvability-related information is represented internally. For the sake of simplicity, in our experiments, for each model, we report results from the layer achieving the highest probing accuracy and use this layer for all subsequent analyses (see ~\ref{sec:appendix_per_layer_probes}). When the optimal layer differs between knowledge and verbalization probes, the performance differences are minor; we therefore select the layer that performs best most consistently within the corresponding model family to facilitate comparison. The selected layers are reported in the experimental settings section. We consider two probing targets. The first is \emph{Knowledge} (K), where probes predict the ground-truth solvability label of the underlying problem. The second is \emph{verbalization} (V), where probes predict whether the model ultimately verbalizes the problem as solvable or unsolvable according to the judge annotations described above. In both cases, we train logistic regression classifiers on hidden states pooled across prompting conditions and evaluate performance on held-out problems.

\begin{table*}[t]
\centering
\begin{tabular}{lccc}
\toprule
& \multicolumn{2}{c}{\textbf{AUC}} & \\ \cmidrule(lr){2-3}
\textbf{Model} & \textbf{Knowledge (K)} & \textbf{Verbalization (V)} & $\mathbf{\cos(K, V)}$ \\
\midrule
LLaMA-3.1-8B-Instruct                    & 0.655 & 0.844 & 0.255 \\
DeepSeek-R1-Distill-LLaMA-8B* & 0.728 & 0.902 & 0.349 \\
Qwen3-4B-Instruct-2507 & 0.790 & 0.831 & 0.473 \\
Qwen3-4B-Thinking-250* & 0.872 & 0.937 & 0.504 \\
Qwen3-30B-A3B-Instruct-2507 & 0.824 & 0.850 & 0.647 \\
Qwen3-30B-A3B-Thinking-2507* & 0.851 & 0.929 & 0.609 \\
Gemma4-31B-Instruct   & 0.814 & 0.884 & 0.380 \\
\bottomrule
\end{tabular}
\caption{Probe performance (ROC-AUC) for solvability knowledge (K) and verbalization (V) on the held-out validation split, aggregated across all prompting conditions. For each model, we report results from the selected layer with the highest probing performance. For reasoning models (*), results are computed using only the output trace.}
\label{tab:probe_auc}
\end{table*}

\paragraph{Representation Steering}

To evaluate the validity of our trained probes and their relation to unfaithful abstention, we perform activation steering on the model's internal hidden states. This allows us to investigate whether explicitly manipulating the encoded features of knowledge and verbalization can successfully rescue abstention failures.

Formally, let $h_t \in \mathbf{R}^d$ represent the hidden state at a targeted layer $L$ and decode token position $t$. We modify the residual stream by adding a steering vector along a specific probe direction:
$$h'_t = h_t + m \cdot \alpha_{\text{unit}} \cdot w$$
where $w \in \mathbf{R}^d$ is the unit-normalized raw direction vector corresponding to a specific probe, and $\alpha_{\text{unit}}$ is the mean residual stream norm at layer $L$. The scalar intensity is controlled by an experimental scaling multiplier $m$.

%------------------------------------------------------------------
\section{Experiments}
\label{sec:experiments}
%------------------------------------------------------------------
\subsection{Experimental Setting}
\paragraph{Datasets:} We conduct our experiments on the ReliableMath benchmark, which consists of 313 solvable problems from widely used mathematical reasoning datasets (MATH \citep{hendrycksmath2021}-high-school level, MinervaMath \citep{mathai_minervamath}-competitive college level, and AIME24 and AMC (\citet{aimo2024a}, \citet{aimo2024b})-Olympiad level) as well as 1,102 unsolvable variants created by removing essential information or introducing contradictory information.
\paragraph{Models:} For our experiments, we query Qwen3-30B-A3B-Instruct-2507, Qwen3-4B-Instruct-2507, Qwen3-30B-A3B-Thinking-2507, Qwen3-4B-Thinking-2507 \citep{yang2025qwen3technicalreport}, Llama-3.1-8B-Instruct \citep{grattafiori2024llama3herdmodels}, DeepSeek-R1-Distill-Llama-8B \citep{Guo_2025} and Gemma4-31B-IT \citep{gemma4_modelcard_2026}. 
\paragraph{Layer Selection:} Layer selection for each model can be found in Table ~\ref{tab:layers}.
\begin{table}[H]
\centering
\label{tab:selected_layers}
\begin{tabular}{lc}
\toprule
\textbf{Model} & \textbf{Layer} \\
\midrule
LLaMA-3.1-8B-Instruct                  & 16 \\
DeepSeek-R1-Distill-LLaMA-8B          & 16 \\
Qwen3-4B-Instruct/Thinking                      & 18 \\
Qwen3-30B-Instruct/Thinking                     & 36 \\
Gemma4-31B-Instruct                    & 36 \\
\bottomrule
\end{tabular}
\caption{Selected layer used for our experiments. For each model, we use the layer with the highest probing performance.}
\label{tab:layers}
\end{table}
\begin{figure*}[t]
\centering
\includegraphics[width=0.75\textwidth]{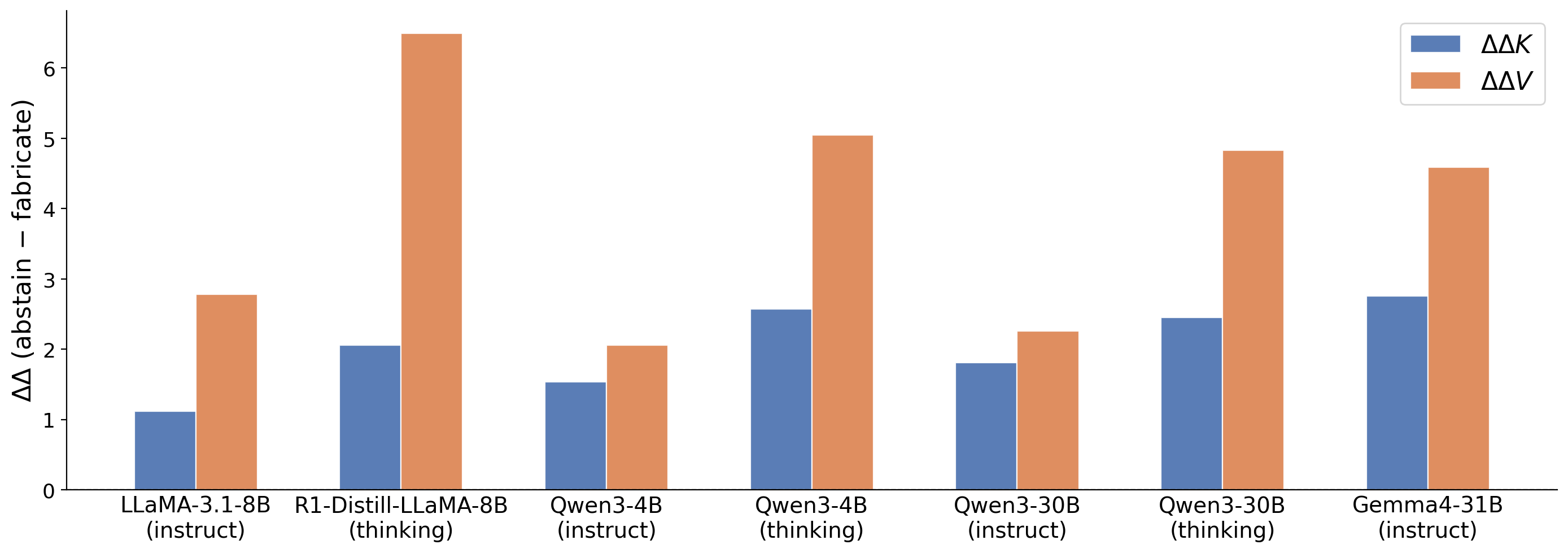}
\caption{Probe sensitivity to activation steering directions for solvability knowledge ($\Delta\Delta K$) and verbalization ($\Delta\Delta V$) on unsolvable problem. The y-axis shows the differential shift ($\Delta\Delta$) between the abstain and fabricate steering conditions on unsolvable problems. Across all evaluated models, verbalization probes consistently exhibit higher sensitivity compared to knowledge probes, especially in reasoning models.}
\label{fig:probe_sensitivity}
\end{figure*}

\subsection{Results}
\subsubsection{RQ1: Knowledge and Verbalization Are Distinct Representations in Model Hidden States}
To assess whether solvability knowledge and solvability verbalization are represented in model hidden states, we evaluate the performance of knowledge and verbalization probes on a held-out validation split across models. Results are shown in Table~\ref{tab:probe_auc} (for the interested reader, results for all probed layers are provided in Appendix~\ref{sec:appendix_per_layer_probes}).
Knowledge and verbalization probes achieve substantially above-chance performance across all models, indicating that both quantities are encoded in model hidden states and can be linearly decoded. Verbalization is consistently easier to decode than knowledge, with V probes achieving higher AUC than K probes for every model. 
%For knowledge, Qwen and Gemma models outperform LLaMA models, with Qwen's reasoning variants achieving the strongest performance. Among the LLaMA models, the reasoning variant also achieves higher knowledge-probe performance, suggesting stronger solvability knowledge encoding in LRMs. For verbalization, the three reasoning models are again the top performers, with the instruction-tuned Gemma 4 31B model ranking second, followed by the instruction-tuned Qwen and Llama models. However, unlike knowledge, verbalization does not exhibit a clear model-family advantage: while Qwen reasoning models achieve the highest overall performance, the ranking among non-reasoning models does not reveal a consistent advantage for either the Qwen or LLaMA family. 
\\
Having established that both quantities are encoded in hidden states, we ask whether they correspond to the same or to distinct directions in hidden-state space. To answer this question, we measure the cosine similarity between the learned knowledge and verbalization probe directions for each model, which is reported in Table ~\ref{tab:probe_auc}. Low similarity indicates that knowledge and verbalization are represented separately.
%The degree of separation varies across models, suggesting that some architectures encode solvability knowledge and its verbalization more independently than others

\subsubsection{RQ2: Fabrication Reflects Knowledge-Verbalization Misalignment Driven Primarily by Verbal Shift}
\paragraph{Misalignment Between Knowledge and Verbalization Is Associated with Fabrication:}
We first measure the correlation between knowledge and verbalization probe activations and compare alignment patterns across fabricated and abstained responses (on the unsolvable-problems' subset). Results are shown in Table~\ref{tab:correlation_fabr_abst}. Across all models, knowledge and verbalization are more strongly aligned during abstention than during fabrication, suggesting that fabrication is associated with a reduction in the alignment between internal solvability knowledge and verbalization. This abstain--fabricate gap is consistently larger for reasoning-oriented variants than for their instruct counterparts, both within Qwen and Llama, indicating that reasoning traces make fabrication-induced misalignment more pronounced across model families. At the same time, the absolute correlation increases with model scale (e.g., Qwen-4B to Qwen-30B), suggesting that larger models maintain stronger overall alignment between knowledge and verbalization, although the fabrication-induced drop persists.

\begin{table}[h]
\centering
\label{tab:r_values}
\begin{tabular}{lcc}
\toprule
\textbf{Model} & $r_{\mathrm{abstain}}$ & $r_{\mathrm{fab}}$ \\
\midrule
LLaMA-3.1-8B-Instruct           & 0.48 & 0.34 \\
DeepSeek-R1-Llama8B*       & 0.47 & 0.25 \\
Qwen3-4B-Instruct               & 0.68 & 0.59 \\
Qwen3-4B-Thinking*   & 0.58 & 0.45 \\
Qwen3-30B-Instruct              & 0.75 & 0.64 \\
Qwen3-30B-Thinking*   & 0.68 & 0.46 \\
Gemma-4-31B-Instruct                 & 0.53 & 0.30 \\
\bottomrule
\end{tabular}
\caption{Classifier correlation $r$ between probe activations and model behaviour under abstain and fabricate steering (unsolvable problems). In all models $r_{\mathrm{abstain}} > r_{\mathrm{fab}}$ (($p < 10^{-4}$, Cohen's $d \in [0.54, 1.10]$)). * denotes response traces.}
\label{tab:correlation_fabr_abst}
\end{table}

\paragraph{Fabrication Primarily Affects Verbalization Rather than Knowledge:}
We next compare the separation between abstained and fabricated responses in the knowledge and verbalization representations (on unsolvable problems). For each model, we compute the difference in probe activations between abstained and fabricated outputs. Results are shown in Figure~\ref{fig:probe_sensitivity}. The abstain--fabricate gap is consistently larger for verbalization than for knowledge, indicating that fabricated responses differ from abstained responses primarily in how solvability is verbalized rather than in the underlying solvability knowledge itself. This is most pronounced in the reasoning models, suggesting that they maintain a more stable representation of solvability knowledge while fabrication arises primarily during the verbalization stage.

\begin{figure*}[t]
\centering
\includegraphics[width=0.8\textwidth]{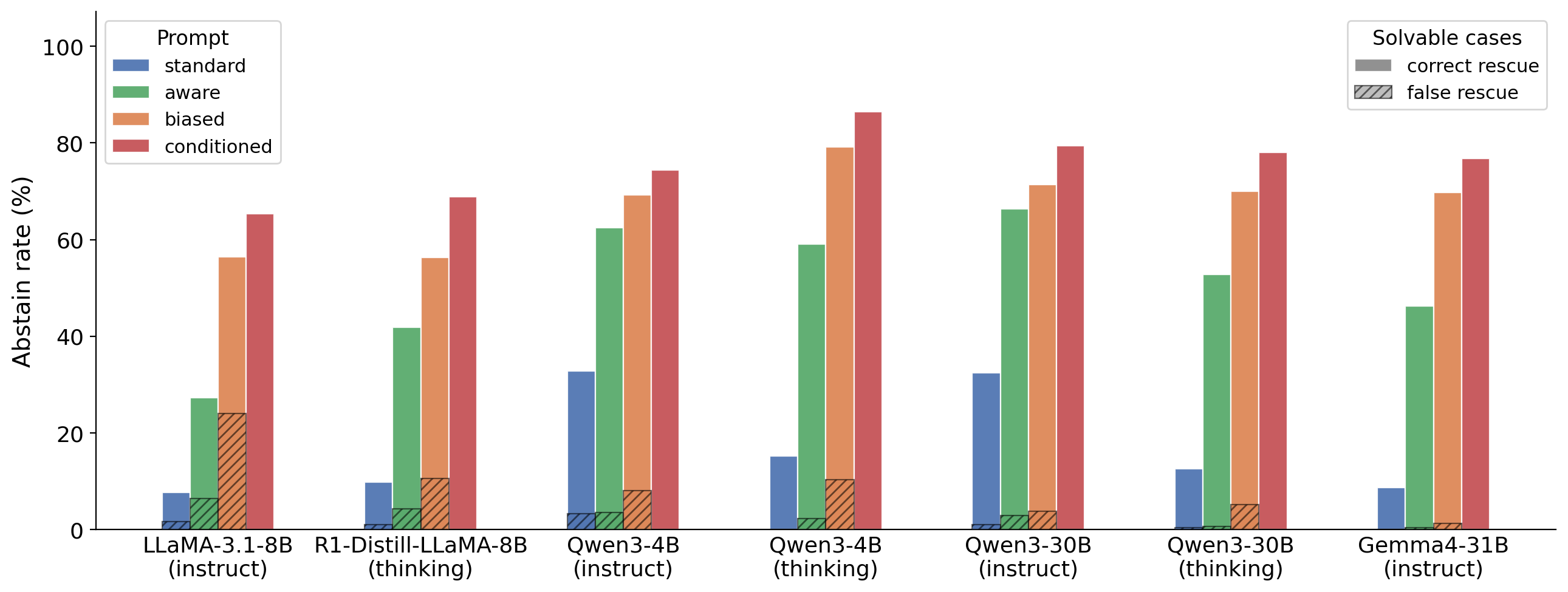}
\caption{Abstention rates on solvable and unsolvable problems under different prompting conditions. Prompts provide progressively stronger cues that a problem may be unsolvable, ranging from \emph{standard} to \emph{conditioned}. Fabrication decreases consistently as prompts become more informative about potential unsolvability.}
\label{fig:prompt_effect}
\end{figure*}

\subsubsection{RQ3: Prompting Reduces Fabrication Primarily Through Changes in Verbalization}
\paragraph{Prompting Increases Abstention Primarily on Unsolvable Problems, Reducing Fabrication:} To understand how prompting affects fabrication, solvability knowledge, and solvability verbalization, we first measure abstention rates on solvable and unsolvable problems under four prompting conditions: \emph{standard}, \emph{aware}, \emph{biased}, and \emph{conditioned} (the prompts are provided in Appendix~\ref{sec:prompts}). These conditions provide progressively stronger signals that a problem may be unsolvable, with the conditioned prompt additionally specifying the source of the unsolvability. Since solvable problems do not contain a source of unsolvability, the conditioned prompt is only applied to unsolvable instances. Results are shown in Figure~\ref{fig:prompt_effect}. Across all models, abstention rates on unsolvable problems increase as prompts provide stronger cues about unsolvability, indicating a substantial reduction in fabrication. A similar trend is observed on solvable problems, but the effect is much weaker: false abstention rates remain consistently low across all prompting conditions. This suggests that prompting can substantially improve recognition of unsolvable instances without inducing widespread over-abstention on solvable ones.

\paragraph{Prompting Affects Verbal Direction, Not Knowledge:} To understand the mechanism behind this improvement, we focus on unsolvable problems, where increased abstention corresponds to correct behavior. We analyze the rate at which the Verbalization (V-only), Knowledge (K-only), and Joint (K and V) probes flip, i.e., the proportion of cases in which a probe's prediction changes under \emph{aware} or \emph{biased} prompting relative to the standard CoT baseline. Table~\ref{tab:probe_flips} presents these probe flip rates across the response traces of the evaluated models.

\begin{table}[h]
    \centering
    % \footnotesize
    \setlength{\tabcolsep}{4.5pt}
    \resizebox{\columnwidth}{!}{
    \begin{tabular}{llccc}
        \toprule
        \textbf{Model} & \textbf{Prompt} & \textbf{V-only} & \textbf{K-only} & \textbf{Joint} \\
        & & \textbf{(\%)} & \textbf{(\%)} & \textbf{(\%)} \\
        \midrule
        \multirow{2}{*}{Qwen3-4B-Instruct} & aware & \textbf{24.0} & 10.2 & 7.3 \\
        & biased & \textbf{29.5} & 8.6 & 11.0 \\
        \midrule
        \multirow{2}{*}{Qwen3-30B-Instruct} & aware & \textbf{23.0} & 10.6 & 8.6 \\
        & biased & \textbf{27.1} & 9.6 & 12.2 \\
        \midrule
        \multirow{2}{*}{Llama-3.1-8B-Instruct} & aware & \textbf{18.9} & 17.6 & 7.7 \\
        & biased & \textbf{35.8} & 13.3 & 17.0 \\
        \midrule
        \multirow{2}{*}{Gemma-4-31B-it} & aware & \textbf{34.6} & 6.6 & 2.0 \\
        & biased & \textbf{55.2} & 8.4 & 6.7 \\
        \midrule
        \multirow{2}{*}{Qwen3-4B-Thinking} & aware & \textbf{33.2} & 11.2 & 10.5 \\
        & biased & \textbf{44.1} & 8.2 & 20.0 \\
        \midrule
        \multirow{2}{*}{Qwen3-30B-Thinking} & aware & \textbf{31.0} & 11.7 & 9.0 \\
        & biased & \textbf{40.0} & 10.0 & 15.7 \\
        \midrule
        \multirow{2}{*}{DeepSeek-R1-Llama8B} & aware & \textbf{27.0} & 8.8 & 5.1 \\
        & biased & \textbf{36.6} & 9.1 & 9.2 \\
        \bottomrule
    \end{tabular}
    }
    \caption{Percentage of probe flips (Verbalization-only, Knowledge-only, Joint) when prompted with \emph{aware} and \emph{biased} instructions compared to standard CoT (unsolvable problems).}
    \label{tab:probe_flips}
\end{table}

%Across all evaluated models, introducing unsolvability cues primarily causes the model's verbalization to flip (resulting in high V-only flip rates), while comparatively few flips occur in the internal Knowledge representations (K-only). Notably, knowledge flips occur less frequently than joint flips (i.e., flips involving both verbal and knowledge components). This suggests that the prompts largely change what the model \emph{says} about solvability, rather than altering its internal \emph{understanding} or knowledge. Moreover, relative to the \emph{aware} prompt, the \emph{biased} prompt consistently increases V-only flips while generally reducing K-only flips, except for slight increases in Gemma 4 and DeepSeek-R1-Llama8B. 
Across all evaluated models, introducing unsolvability cues primarily causes the model's verbalization to flip (resulting in high V-only flip rates), while comparatively few flips occur in the internal knowledge representations (K-only), with this disparity generally becoming more pronounced under the \emph{biased} prompt than under the \emph{aware} prompt. Notably, K-only flips occur less frequently than joint flips (i.e., simultaneous flips in both knowledge and verbal representations). This suggests that the prompts primarily change what the model \emph{says} about solvability, rather than its internal \emph{understanding}.
%This further indicates that stronger unsolvability cues primarily influence verbalization rather than the underlying representation of solvability. 
The V-only vs K-only difference is statistically significant in every row (two-proportion $z$-test, $p < 10^{-5}$), with the exception of Llama-3.1-8B/aware. The same patterns hold for the conditioned prompt; results are provided in the appendix ~\ref{sec:appendix_conditioned_flip} due to space constraints.
% We further assess whether prompting alters the strength with which these quantities are encoded in hidden states by comparing knowledge and verbalization probe performance across prompting conditions. Results are shown in Figure~\ref{fig}. [Results interpretation.]

\subsubsection{RQ4: Reasoning and Output Traces of LRMs handle Knowledge and Verbalization differently}

We next examine how Knowledge and Verbalization differ between the thinking and output traces of LRMs. 
\paragraph{Knowledge and Verbalization Are Differentially Affected by Prompt Bias in Reasoning and Output Traces:} First, we investigate how susceptible each type of trace is to prompts containing progressively stronger unsolvability cues. To do so, we examine the distribution of Knowledge and Verbalization probe flips on unsolvable problems. Results can be found in table~\ref{tab:trace_comparison}.

\begin{table}[h]
    \centering
    % \footnotesize
    \setlength{\tabcolsep}{3.5pt} % Tightened to fit in a single column
    \resizebox{\columnwidth}{!}{
    \begin{tabular}{lllccc}
        \toprule
        \textbf{Model} & \textbf{Prompt} & \textbf{Trace} & \textbf{V-only} & \textbf{K-only} & \textbf{Joint} \\
        & & & \textbf{(\%)} & \textbf{(\%)} & \textbf{(\%)} \\
        \midrule
        \multirow{4}{*}{Qwen3-4B-T} & \multirow{2}{*}{aware} 
        & Thinking & 17.7 & 10.0 & 5.4 \\
        & & Response & \textbf{33.2} & \textbf{11.2} & 10.5 \\
        \cmidrule{2-6}
        & \multirow{2}{*}{biased} 
        & Thinking & 25.5 & \textbf{9.1} & 8.1 \\
        & & Response & \textbf{44.1} & 8.2 & 20.0 \\
        \midrule
        \multirow{4}{*}{Qwen3-30B-T} & \multirow{2}{*}{aware} 
        & Thinking & 19.7 & \textbf{12.3} & 5.1 \\
        & & Response & \textbf{31.0} & 11.7 & 9.0 \\
        \cmidrule{2-6}
        & \multirow{2}{*}{biased} 
        & Thinking & 26.6 & \textbf{13.1} & 10.1 \\
        & & Response & \textbf{40.0} & 10.0 & 15.7 \\
        \midrule
        \multirow{4}{*}{R1-Llama8B} & \multirow{2}{*}{aware} 
        & Thinking & 13.0 & \textbf{11.8} & 5.2 \\
        & & Response & \textbf{27.0} & 8.8 & 5.1 \\
        \cmidrule{2-6}
        & \multirow{2}{*}{biased} 
        & Thinking & 16.7 & \textbf{10.8} & 6.4 \\
        & & Response & \textbf{36.6} & 9.1 & 9.2 \\
        \bottomrule
    \end{tabular}
    }
    \caption{Comparison of probe flip rates between the internal \emph{thinking} trace and the final \emph{response} trace under \emph{aware} and \emph{biased} prompts for unsolvable problems. (R1-Llama8B denotes DeepSeek-R1-Llama8B).}
    \label{tab:trace_comparison}
\end{table}

Across all models and prompt conditions, both reasoning- and response-level V-flips are more pronounced under the biased prompt than under the aware prompt, reinforcing the hypothesis that biased prompts primarily alter what models verbalize about solvability rather than their underlying solvability knowledge. Furthermore, comparing the internal reasoning traces and final response traces of LRMs reveals contrasting patterns for K-only and V-only flips. Specifically, V-only flip rates are substantially lower in thinking traces than in final responses, suggesting that the response generation stage is particularly susceptible to surface-level shifts in verbalized solvability judgments. In contrast, although K-only flips remain rare overall, they occur more frequently in thinking traces than in final responses across most models and prompt conditions, indicating that perturbations along the knowledge direction are more readily reflected in the model's internal reasoning process than in its final output (see appendix ~\ref{sec:appendix_conditioned_flip} for conditioned prompt results). All V-only vs K-only differences in Table~\ref{tab:trace_comparison} are statistically significant ($p < 10^{-4}$, two-proportion $z$-test), except DeepSeek-R1-Llama8B/aware.

\begin{figure}[h]
    \centering
    \includegraphics[width=\columnwidth]{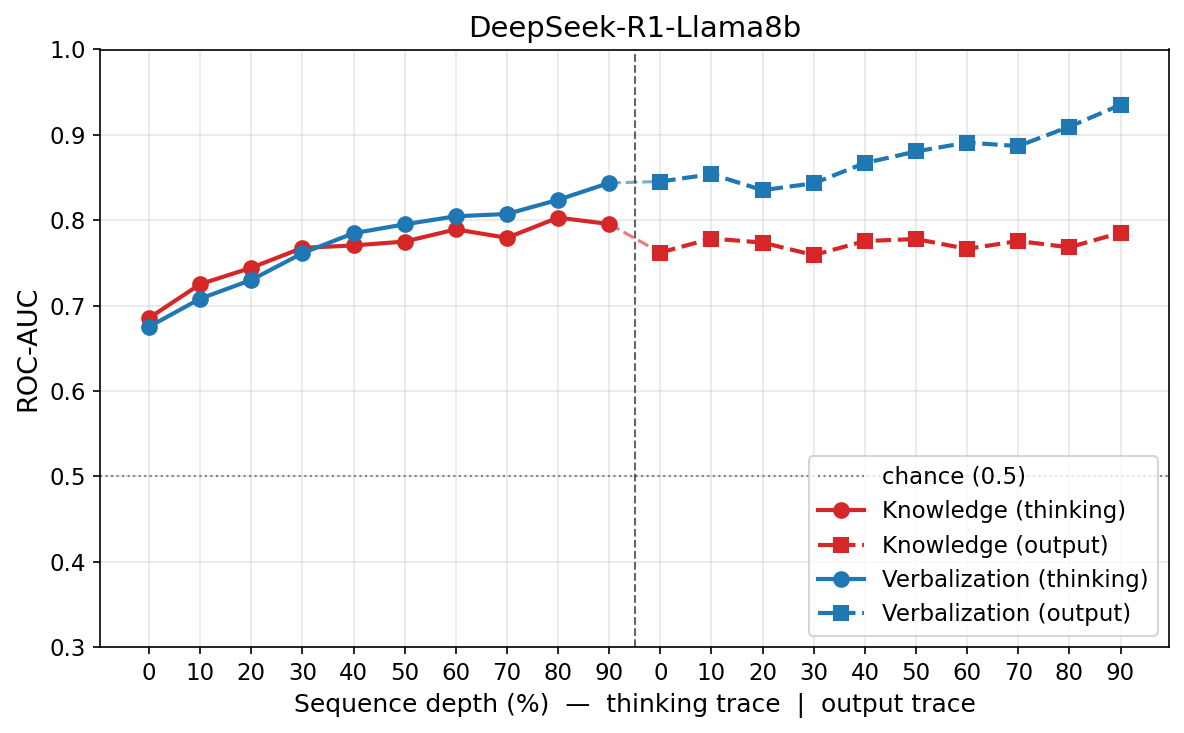}
    \caption{Knowledge- and verbalization-probe ROC-AUC across sequence-depth bins of the thinking trace (left of the dashed line) and the output trace (right) for DeepSeek-R1-Llama8B.}
    \label{fig:depth_llama}
\end{figure}

\paragraph{Thinking and Output Traces have Distinct Depth-Wise Dynamics of Knowledge and Verbalization:} We next investigate the depth-wise behavior of the knowledge and verbalization representations in Figure~\ref{fig:depth_llama}, which plots knowledge- and verbalization-probe ROC-AUC across sequence-depth bins of the thinking trace (left of the dashed line) and the output trace (right). Due to space limits, we focus on DeepSeek-R1-Llama8B; the same patterns hold across all reasoning models (details in Appendix ~\ref{sec:appendix_depth}). Two trends emerge. First, knowledge is decodable above chance from the earliest bins and remains comparatively flat throughout and slightly declines across the output trace-indicating that the model's latent solvability signal is established early and is largely trace-invariant. Second, verbalization starts below knowledge in the early thinking bins but rises monotonically with depth, overtaking knowledge partway through and peaking in the output trace. The verbalized verdict therefore consolidates gradually over generation and is most strongly encoded once the model commits to a final answer. Finally, the K-V gap widens through the output trace.

%We then compare the effect of prompting on the underlying knowledge and verbalization representations. For each trace type, we measure the prompt-induced changes in knowledge and verbalization probe outputs. Results are shown in Figure~\ref{fig}. [Results interpretation.]

\subsubsection{RQ5. Knowledge, Verbalization, and Their Alignment Can Be Selectively Steered}
\label{subsec:rq5}

To evaluate whether selectively intervening along these latent representations can rescue model abstention, we experiment with steering along three directions: the knowledge direction (K, from the Knowledge probe), the verbalization direction (V, from the Verbalization probe), and their normalized combination (KV). Steering along KV pushes knowledge and verbalization jointly rather than independently, aligning with the intuitive hypothesis that fabrication stems from a misalignment between what the model knows and what it verbalizes. Under this hypothesis, acting on both directions simultaneously is more likely to restore their agreement than steering either direction alone.

We apply activation steering to 280-sample subsets from each of our two problem splits: unsolvable problems (the rescue target) and solvable problems (to monitor specificity and false-alarm side effects). We sweep the steering multiplier m and report the resulting abstention rate. Our results reveal a performance–specificity trade-off governed by the gating criterion, which we present in two parts. Due to space constraints, we report results for Qwen3-4B-Instruct; (results for the other models are in Appendix~\ref{sec:appendix_steering})

\begin{figure}[h]
    \centering
    \includegraphics[width=0.45\textwidth]{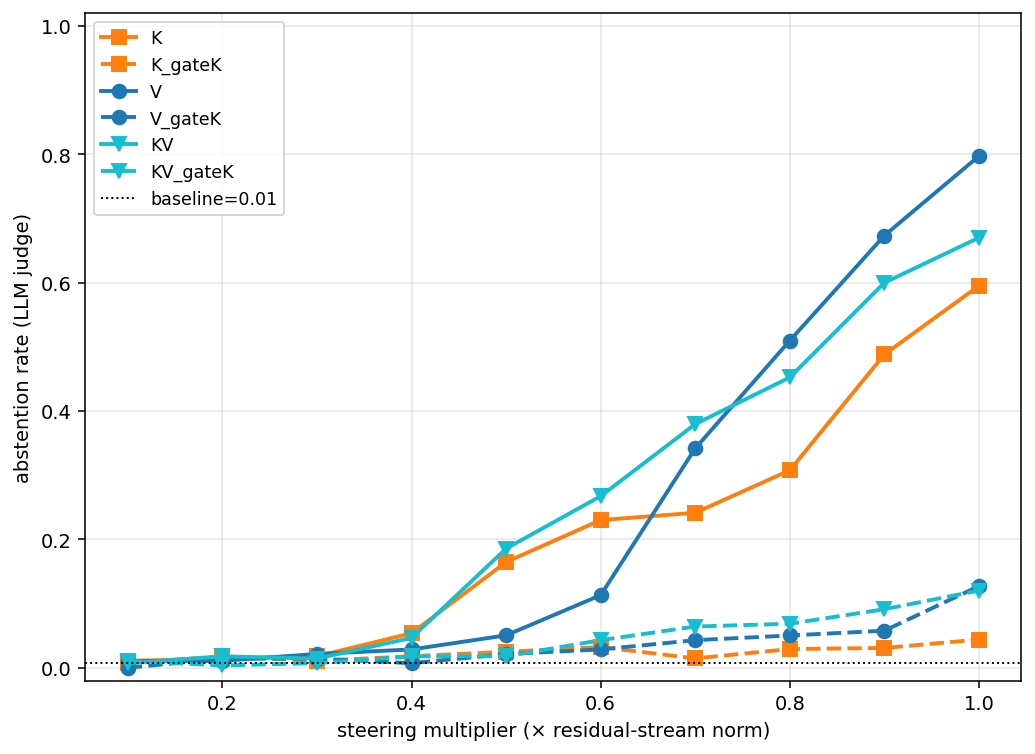}
    \caption{Steering dose-response on the \textbf{solvable} split (specificity control) for Qwen3-4b-instruct, layer~18; lower abstention is better
    }
    \label{fig:steer_solvable}
\end{figure}

\paragraph{Ungated steering is catastrophic for specificity} On the solvable control split (Figure~\ref{fig:steer_solvable}), static, non-gated interventions along the pure V, pure K, and combined KV directions  are severely destructive: as the multiplier increases, they drive abstention on correctly solvable problems far above the 0.01 baseline, reaching roughly 0.6–0.8 at high intensities for the V and KV directions. In other words, ungated steering causes the model to abandon problems it could otherwise solve, fabricating spurious abstentions on a large fraction of benign instances.

 To overcome this, we introduce a discrete, binary token-level gating mechanism conditioned on the model's token-level internal states  . Under this policy, steering is only activated at token positions where the Knowledge probe's predicted probability $p_K(h_t)$ exceeds a confidence threshold of $0.5$:
$$g_t = \begin{cases} m & \text{if } p_K(h_t) > 0.5 \\ 0 & \text{otherwise} \end{cases}$$

By dynamically restricting steering strictly to tokens where the model demonstrates latent solving capacity, this gating approach drastically minimizes false positives (incorrectly altering solvable problems) while maintaining a high rate of successful faithful rescues.

\begin{figure}[h]
    \centering
    \includegraphics[width=0.45\textwidth]{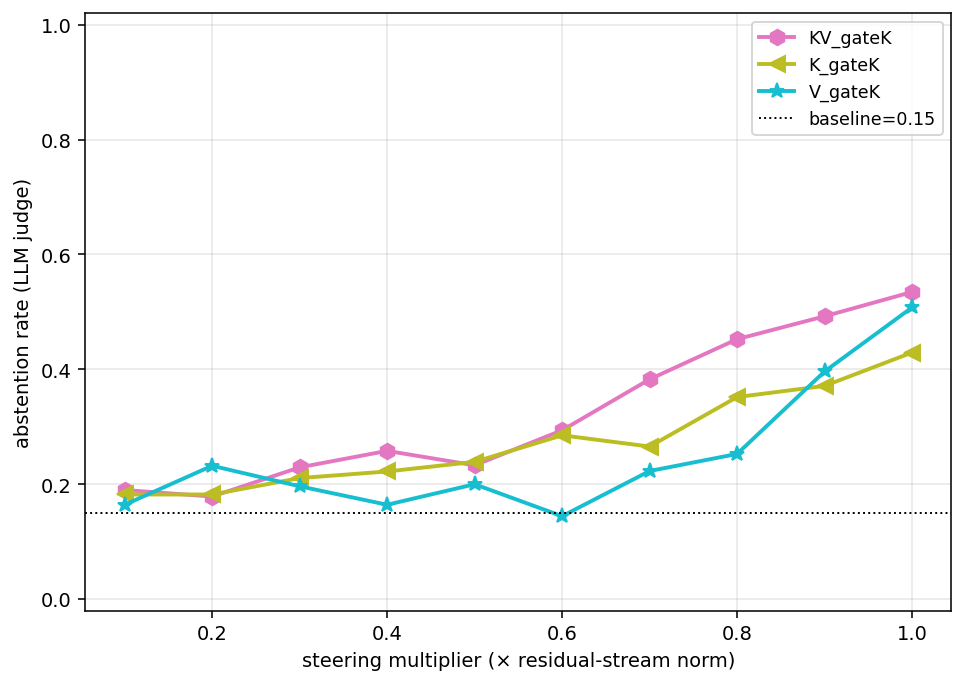}
    \caption{Steering dose-response on the \textbf{unsolvable} split for Qwen3-4b-instruct, layer~18; higher abstention is better.}
    \label{fig:steer_unsolvable}
\end{figure}

\paragraph{Gated KV outperforms gated K or V} Having established that gating is necessary, Figure~\ref{fig:steer_unsolvable} reports the three gated configurations on the unsolvable target split, where higher abstention indicates successful rescue of an otherwise-fabricated response. KVgateK is the strongest and most stable, sitting at or above both single-direction variants, reaching the highest rescue rate at maximum intensity. Steering the combined direction is thus more reliable than steering either alone. This holds for Qwen3-4B, where K and V are relatively misaligned directions (cosine similarity $= 0.47$); see Appendix~\ref{sec:appendix_steering} for other models.

%This behavior causally proves that the alignment of latent knowledge and verbalization traits can be controlled selectively. Rather than a blunt, global vector injection, surgically triggering interventions on structural knowledge availability successfully mitigates unfaithful abstentions without corrupting the model's underlying performance.

\section{Conclusions}
We investigate the internal representations of solvability knowledge and verbalization in LLMs and show that they are encoded as distinct, linearly decodable directions, with verbalization consistently easier to decode than knowledge. Fabrication is associated with a reduced alignment between these representations and primarily reflects changes in verbalization rather than the underlying knowledge. Prompting with unsolvability cues reduces fabrication mainly by shifting the verbalization representation, while reasoning and response traces exhibit distinct dynamics under prompt bias. Finally, although ungated steering severely compromises specificity, gated joint steering of knowledge and verbalization consistently outperforms steering either direction alone, yielding the most reliable improvements in abstention.

\section*{Limitations}
%We acknowledge that our work is restricted to mathematical solvability, and it remains an open question whether the distinction between knowledge and verbalization generalizes to broader answerability and hallucination settings. Investigating this generalization is an important direction for future work.
We acknowledge that our analysis focuses on the interplay between the internal representations of knowledge and verbalization, rather than on the more traditional notion of faithfulness, which examines the relationship between internal representations and generated text. We leave bridging these two perspectives to future work. Additionally, our study is restricted to mathematical solvability, and it remains an open question whether the distinction between knowledge and verbalization extends to general question answering and other settings. Investigating the generality of this distinction across such domains is an important direction for future research.

% Bibliography entries for the entire Anthology, followed by custom entries
%\bibliography{anthology,custom}
% Custom bibliography entries only
\bibliography{custom}

\appendix

\section{Prompts}
\label{sec:prompts}
In this section, we provide the prompts used in our pipeline: the CoT generation prompts (Table ~\ref{tab:cot_prompts}) and the LLM-as-a-judge prompt for verbalization labeling (Table ~\ref{tab:judge_prompt}).

\section{Details on Data Preparation for Probing}
\label{sec:appendix_data_prep}

The data that was used for training our probes consisted of approximately 100k training samples and 20k validation samples for the solvability probe (K), and approximately 90k training and 18k validation samples for the verbalization probe (V). The solvability probe is trained on a class distribution of ~81\% unsolvable vs. ~19\% solvable, reflecting the dataset construction. The verbalization probe class balance varies by model, ranging from 23\% to 46\% positive (i.e., the model verbalized unsolvability), capturing genuine differences in model behavior.

\section{Problem and Responses Examples}
\label{sec:appendix_examples}

We document two examples from our dataset to illustrate how models navigate unsolvable problems. Table \ref{tab:example_fabrication} provides a detailed, full-text look at the fabrication behavior previously illustrated in Figure \ref{fig:overview}, where the model's internal knowledge recognizes the unsolvability, but its verbalization fails to express it. Table \ref{tab:example_abstention} showcases a successful abstention where internal knowledge and verbalization align.

\section{Per Layer Linear Probes AUC}
\label{sec:appendix_per_layer_probes}

\begin{figure*}[t]
    \centering
    
    % --- Top Left: LLaMA Pair ---
    \begin{subfigure}[b]{0.48\textwidth}
        \centering
        \includegraphics[width=\textwidth]{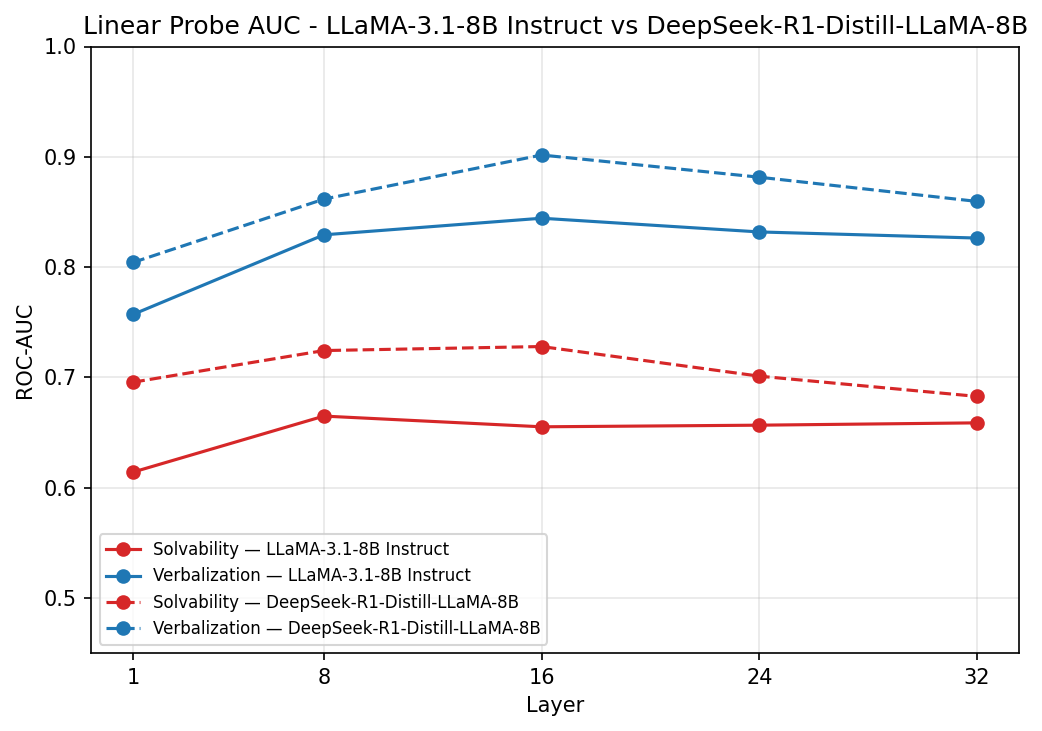} % Replace with actual filename
        \caption{LLaMA-3.1-8B-Instruct vs DeepSeek-R1-Distill-LLaMA-8B}
        \label{fig:layer_llama}
    \end{subfigure}
    \hfill
    % --- Top Right: Qwen-4B Pair ---
    \begin{subfigure}[b]{0.48\textwidth}
        \centering
        \includegraphics[width=\textwidth]{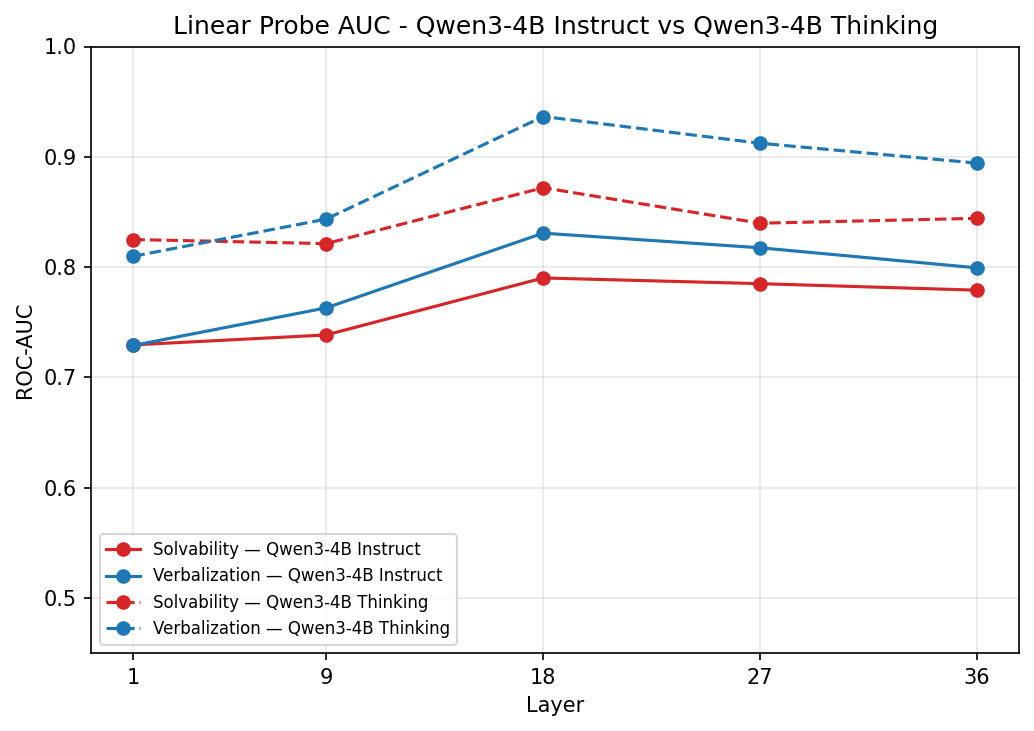} % Replace with actual filename
        \caption{Qwen3-4B-Instruct vs Qwen3-4B-Thinking}
        \label{fig:layer_qwen4b}
    \end{subfigure}
    
    \vspace{0.4cm} % Vertical space between rows
    
    % --- Bottom Left: Qwen-30B Pair ---
    \begin{subfigure}[b]{0.48\textwidth}
        \centering
        \includegraphics[width=\textwidth]{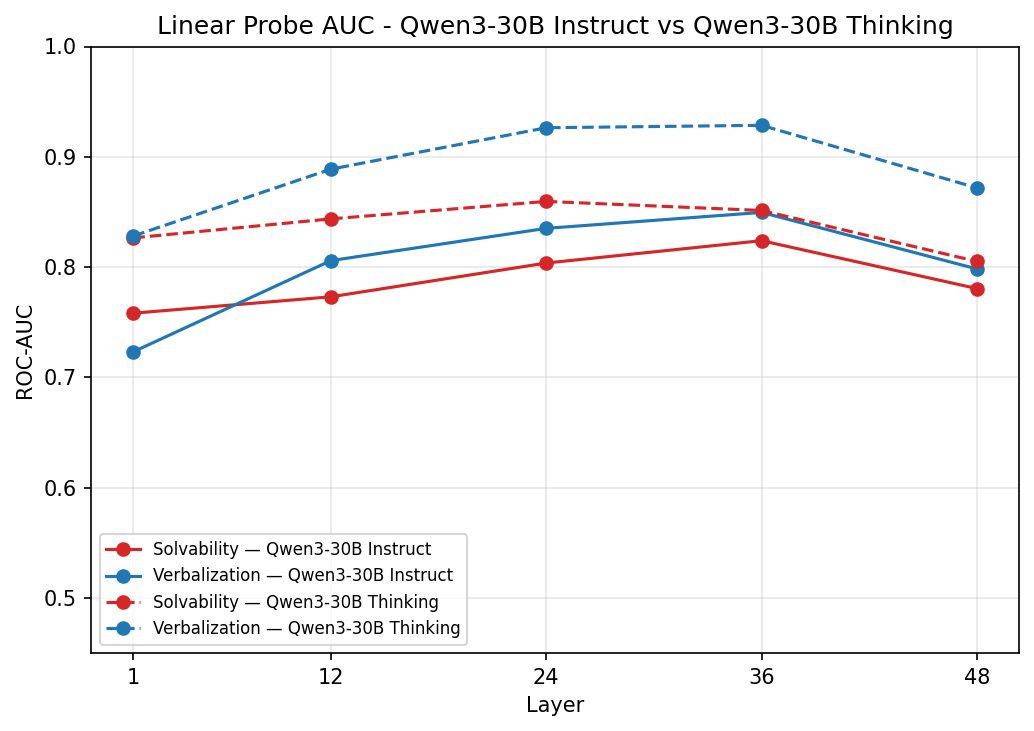} % Replace with actual filename
        \caption{Qwen3-30B-Instruct vs Qwen3-30B-Thinking}
        \label{fig:layer_qwen30b}
    \end{subfigure}
    \hfill
    % --- Bottom Right: Gemma ---
    \begin{subfigure}[b]{0.48\textwidth}
        \centering
        \includegraphics[width=\textwidth]{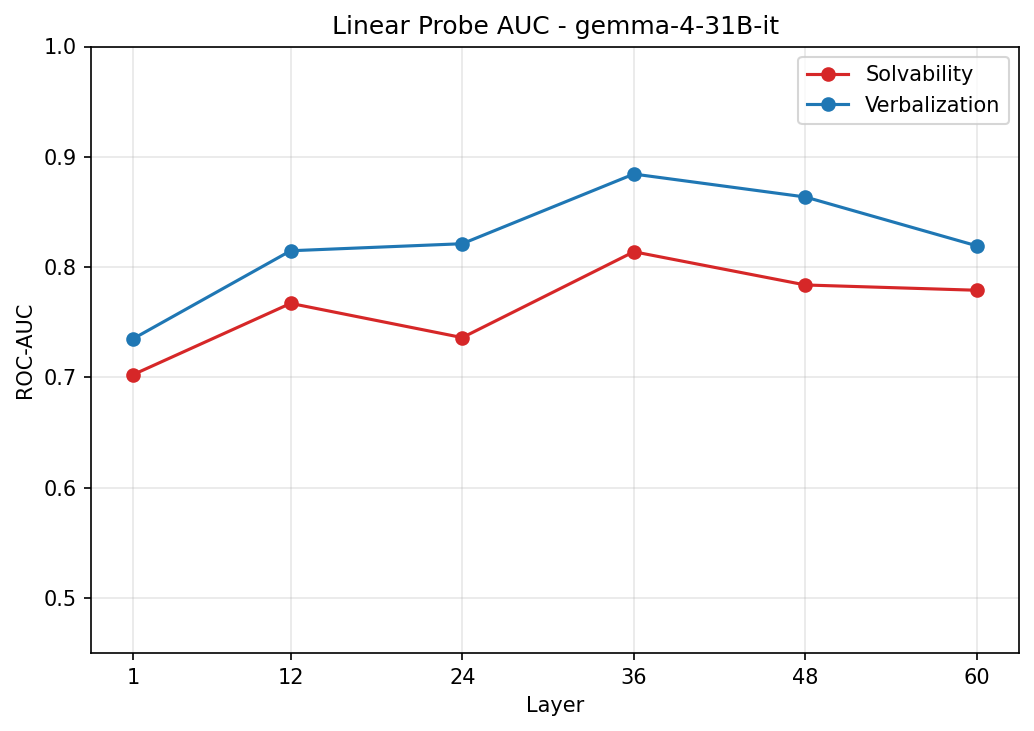} % Replace with actual filename
        \caption{Gemma4-31B-Instruct}
        \label{fig:layer_gemma}
    \end{subfigure}
    
    \caption{Per-layer linear probe ROC-AUC for knowledge (solvability) and verbalization targets. The plots illustrate the performance trajectories across network depth for Instruct versus Thinking/Reasoning models, demonstrating that peak separability generally occurs in the middle-to-late layers.}
    \label{fig:per_layer_auc}
\end{figure*}

In the main text, we report probing results using the hidden states from the layer that achieved the highest accuracy for each respective model (as outlined in Table~\ref{tab:layers}). To provide a comprehensive view of how these internal representations evolve across the network depth, Figure~\ref{fig:per_layer_auc} presents the linear probe ROC-AUC across all sampled layers. 

We compare the layer-wise trajectories of the Instruct and Thinking/Reasoning variants for the LLaMA-8B, Qwen-4B, and Qwen-30B model families, alongside the Gemma-31B-Instruct model. Across most models, we observe a consistent trajectory: probing performance for both knowledge (solvability) and verbalization typically peaks in the middle-to-late layers (e.g., layer 16 for LLaMA-8B and layer 18 for Qwen-4B) before plateauing or slightly degrading in the final layers. Furthermore, the Thinking and Reasoning variants consistently exhibit stronger class separability (higher AUC) than their base Instruct counterparts across the network depth. These trajectories motivate and justify the optimal layer selections used in our primary analyses.

\section{Human Validation of LLM-as-a-judge}
\label{sec:appendix_human_annotation}
To validate the reliability of the LLM-as-a-judge, two of the authors manually annotated a random subset of 100 reasoning traces according to the same criterion (and based on the prompt) used by the judge: whether the model's dominant behavior throughout the reasoning process was to attempt to solve the problem or to conclude that the problem is unsolvable. Table~\ref{tab:judge_validation} reports the agreement between the human annotations and the LLM judge.

\begin{table}[h]
\centering
\small
\begin{tabular}{lc}
\toprule
\textbf{Annotator Pair} & \textbf{Cohen's $\kappa$} \\
\midrule
LLM -- Human 1 & 0.9400 \\
LLM -- Human 2 & 0.9400 \\
Human 1 -- Human 2 & 0.9600 \\
\bottomrule
\end{tabular}

\caption{Inter-annotator agreement on a manually annotated subset of 100 reasoning traces. Each trace was labeled according to its dominant behavior: whether the model primarily attempts to solve the problem or concludes that it is unsolvable. The high Cohen's $\kappa$ values indicate strong agreement between the LLM judge and the human annotators, comparable to the agreement between the two human annotators.}
\label{tab:judge_validation}
\end{table}

\begin{table*}[t]
\centering
\small
\begin{tabular}{p{0.20\textwidth} p{0.74\textwidth}}
\toprule
\textbf{Variant} & \textbf{Prompt} \\
\midrule

Standard &
\parbox[t]{\linewidth}{\ttfamily
Solve the following math problem step by step. Put your final answer within \textbackslash boxed\{\}\\. Problem: \{question\}
} \\
\midrule

Standard plus Solvability &
\parbox[t]{\linewidth}{\ttfamily
Solve the following math problem step by step.\\
Put your numerical answer within \textbackslash boxed\{\}.\\
Then, on a new line, state whether the problem is solvable or unsolvable:\\
- \textbackslash boxed\{The problem is solvable\}\\
- \textbackslash boxed\{The problem is unsolvable\}\\
Problem: \{question\}
} \\
\midrule

Unsolvable Aware &
\parbox[t]{\linewidth}{\ttfamily
Solve the following math problem step by step, or state that it is unsolvable.\\
Put your final answer in exactly one of these formats:\\
- If the problem is solvable: \textbackslash boxed\{<answer>\}\\
- If the problem cannot be solved with the information provided: \textbackslash boxed\{unsolvable\}\\
Problem: \{question\}
} \\
\midrule

Unsolvability Biased &
\parbox[t]{\linewidth}{\ttfamily
Solve the following math problem step by step, or state that it is unsolvable.\\
Put your final answer in exactly one of these formats:\\
- If the problem is solvable: \textbackslash boxed\{<answer>\}\\
- If the problem cannot be solved with the information provided: \textbackslash boxed\{unsolvable\}\\
It is highly likely that the problem is unsolvable.\\
Problem: \{question\}
} \\
\midrule

Conditioned &
\parbox[t]{\linewidth}{\ttfamily
Solve the following math problem step by step, or state that it is unsolvable.\\
Put your final answer in exactly one of these formats:\\
- If the problem is solvable: \textbackslash boxed\{\textless answer\textgreater\}\\
- If the problem cannot be solved with the information provided: \textbackslash boxed\{unsolvable\}\\
It is highly likely that the problem is unsolvable due to [A]/[B].\\
\textbf{[A] contradiction:} The problem contains two or more statements that cannot all be true at the same time, so no valid solution can be determined.\\
\textbf{[B] removal:} The problem is missing critical information needed to solve it, so there is not enough information to determine an answer.\\
Problem: \{question\}}
\\
\bottomrule
\end{tabular}
\caption{Prompts used for CoT generation.}
\label{tab:cot_prompts}
\end{table*}

\begin{table*}[t]
\centering
\small
\begin{tabular}{p{0.94\textwidth}}
\toprule
\textbf{Judge Prompt} \\
\midrule

\ttfamily
You are an annotator. A solver is given a problem and produces a response. The problem may be solvable or unsolvable. Your task is to determine how the solver ultimately treated the problem.\\[0.5em]

Classification rules:\\[0.25em]

"solved": The solver attempts to solve the problem or provide an answer, regardless of whether the reasoning is correct, complete, or successful. This includes:\\
- Partial solutions\\
- Incomplete calculations\\
- Incorrect reasoning\\
- Failed attempts that still try to derive an answer\\
- Any response that does not ultimately conclude the problem is unsolvable\\[0.5em]

"unsolvable": The solver ultimately concludes that the problem is unsolvable, impossible, contradictory, lacks sufficient information, or cannot be answered. If the solver reaches this conclusion and does not later return to solving the problem, classify it as "unsolvable".\\[0.5em]

Important notes:\\
- Solvers may change their minds during the response. Base your judgment on their final position.\\
- Do not evaluate whether the solver's reasoning or answer is correct.\\
- Do not determine whether the problem is actually solvable.\\
- Only determine whether the solver ultimately treated the problem as solvable (and attempted to solve it) or as unsolvable.\\[0.5em]

Always return valid JSON with exactly these two fields:\\
\{\\
\ \ "verbal\_annotation": "solved" | "unsolvable",\\
\ \ "judgement\_explanation": "brief explanation"\\
\}

\\
\bottomrule
\end{tabular}
\caption{Prompt used to classify whether a model ultimately treated a problem as solved or unsolvable.}
\label{tab:judge_prompt}
\end{table*}

\begin{table*}[t]
\centering
\small
\begin{tabular}{p{0.94\textwidth}}
\toprule
\textbf{Example 1: Fabrication} \\
\midrule

\textbf{Features:}\\
$\bullet$ \textbf{Model:} \texttt{llama\_31\_8b\_instruct} \quad $\bullet$ \textbf{Layer:} 16 \quad $\bullet$ \textbf{Prompt:} standard\\
$\bullet$ \textbf{Split:} unsolvable (ground truth) \quad $\bullet$ \textbf{Edit type:} contradict\\
$\bullet$ \textbf{Judge label:} solved (model fabricated an answer)\\
$\bullet$ \textbf{K probe score:} +0.72 (predicts unsolvable)\\
$\bullet$ \textbf{V probe score:} -2.59 (predicts solved)\\
$\bullet$ \textbf{Predicted answer:} 1170\\[0.5em]

\textbf{Input Problem:}\\
How many license plates can be formed if every license plate has 2 different letters (A through Z) followed by 2 different one digit (0-9) numbers, and the two letters must be the same?\\[0.5em]

\textbf{Model Response:}\\
To solve this problem, we need to break it down into steps:\\[0.25em]
Choose the first letter: Since there are 26 letters in the alphabet and we want the two letters to be the same, we have 26 choices for the first letter.\\[0.25em]
Choose the second letter: Since the second letter must be the same as the first letter, there is only 1 choice for the second letter.\\[0.25em]
Choose the first digit: There are 10 choices for the first digit (0-9).\\[0.25em]
Choose the second digit: Since the second digit must be different from the first digit, there are 9 choices for the second digit.\\[0.25em]
Calculate the total number of license plates: Multiply the number of choices for each step together.\\[0.25em]
So, the total number of license plates is: $26 \times 1 \times 10 \times 9 = 2340$.\\[0.25em]
However, we have counted each license plate twice because for each pair of letters, the order of the letters doesn't matter. To correct this, we divide the total by 2.\\[0.25em]
So, the total number of license plates is: $2340 / 2 = 1170$.\\[0.25em]
Therefore, the final answer is: $\boxed{1170}$

\\
\bottomrule
\end{tabular}
\caption{Example of fabrication where the model's internal knowledge recognizes the unsolvability, but its verbalization fails to express it.}
\label{tab:example_fabrication}
\end{table*}

\begin{table*}[t]
\centering
\small
\begin{tabular}{p{0.94\textwidth}}
\toprule
\textbf{Example 2: Successful Abstention} \\
\midrule

\textbf{Features:}\\
$\bullet$ \textbf{Model:} \texttt{llama\_31\_8b\_instruct} \quad $\bullet$ \textbf{Layer:} 16 \quad $\bullet$ \textbf{Prompt:} standard\\
$\bullet$ \textbf{Split:} unsolvable (ground truth) \quad $\bullet$ \textbf{Edit type:} contradict\\
$\bullet$ \textbf{Judge label:} unsolvable (model correctly abstained)\\
$\bullet$ \textbf{K probe score:} +2.55 (predicts unsolvable)\\
$\bullet$ \textbf{V probe score:} +1.92 (predicts unsolvable)\\
$\bullet$ \textbf{Predicted answer:} undefined\\[0.5em]

\textbf{Input Problem:}\\
Remmy wants to divide 10 by 0, but he cannot remember how to do that. By what number should he multiply 10 to get the answer?\\[0.5em]

\textbf{Model Response:}\\
To solve this problem, we need to understand that division and multiplication are inverse operations, meaning that dividing a number by another number is the same as multiplying that number by the reciprocal of the other number.\\[0.25em]
So, if we want to divide 10 by a number, we need to multiply 10 by the reciprocal of that number. The reciprocal of 0 is undefined, as division by zero is not allowed in mathematics.\\[0.25em]
However, the problem asks what number Remmy should multiply 10 by to get the answer. Since division by zero is undefined, there is no number that Remmy can multiply 10 by to get the answer. In other words, the answer is undefined.\\[0.25em]
Therefore, the answer is $\boxed{\text{undefined}}$

\\
\bottomrule
\end{tabular}

\caption{Example of a successful abstention where internal knowledge and verbalization align.}
\label{tab:example_abstention}
\end{table*}

\section{Flips under the Conditioned Prompt}
\label{sec:appendix_conditioned_flip}
Table~\ref{tab:conditioned_flip} reports the percentage of probe flips (Verbalization-only, Knowledge-only, and Joint) under the conditioned prompt relative to the standard CoT baseline on unsolvable problems. We omit these results from the main text due to space constraints. Overall, they exhibit the same pattern as those observed for the aware and biased prompts.

\section{Depth-Wise Dynamics of Knowledge and Verbalization: Additional Models}
\label{sec:appendix_depth}

\begin{figure*}[t]
    \centering
    \begin{subfigure}[b]{0.48\textwidth}
        \centering
        \includegraphics[width=\textwidth]{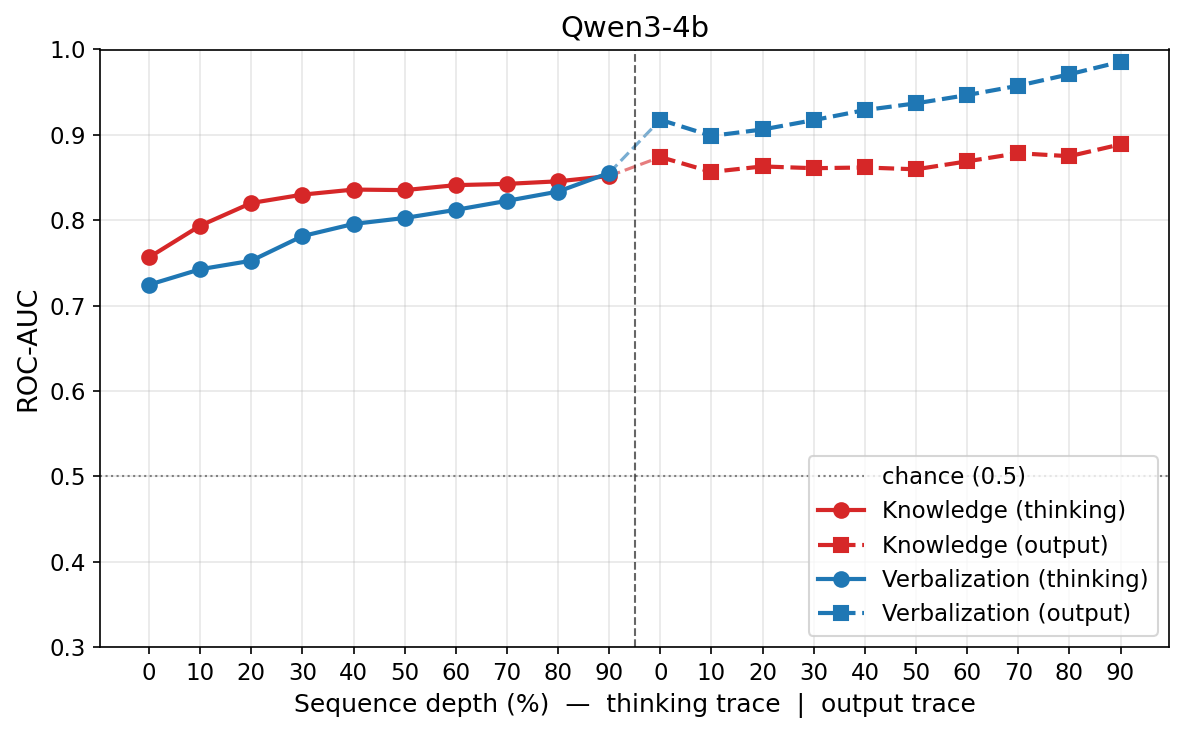}
        \caption{Qwen-4B-Thinking}
        \label{fig:qwen_4b}
    \end{subfigure}
    \hfill
    \begin{subfigure}[b]{0.48\textwidth}
        \centering
        \includegraphics[width=\textwidth]{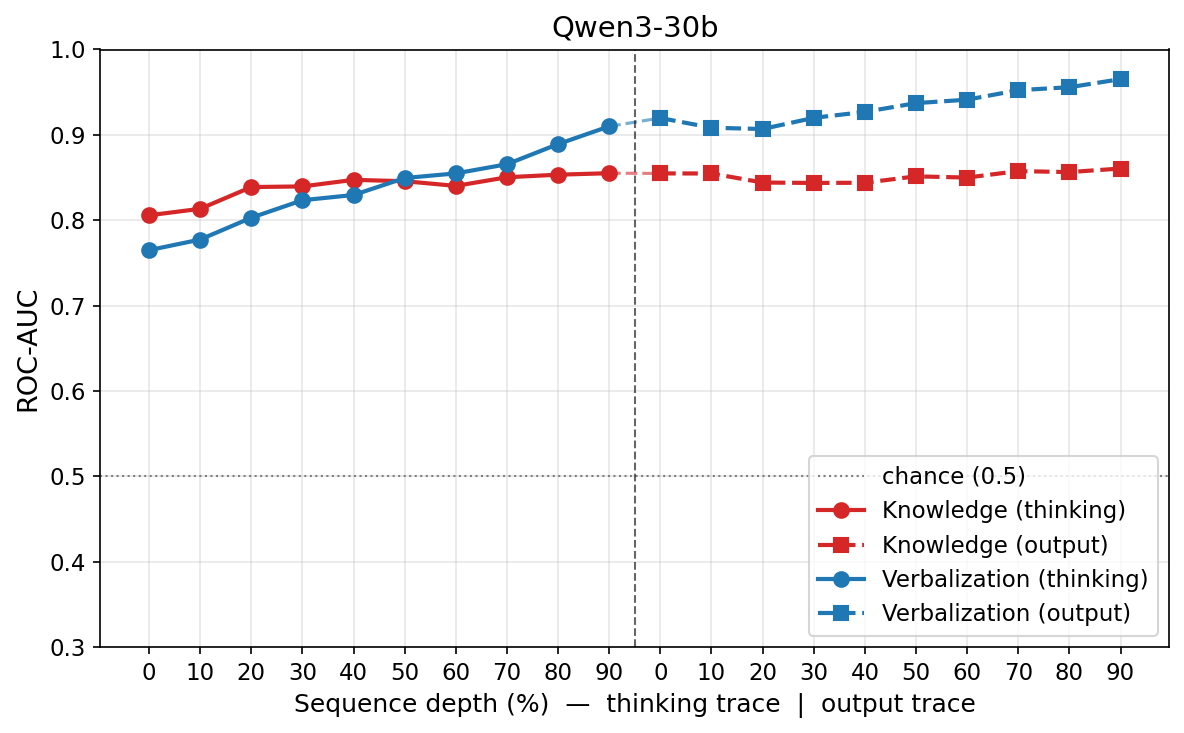}
        \caption{Qwen-30B-Thinking}
        \label{fig:qwen_30b}
    \end{subfigure}
    \caption{Depth-wise dynamics of knowledge and verbalization for Qwen-4B-Thinking and Qwen-30B-Thinking.}
    \label{fig:depth_dynamics}
\end{figure*}

In Figure~\ref{fig:depth_dynamics}, we provide the depth-wise knowledge and verbalization plots for Qwen-4B-Thinking and Qwen-30B-Thinking, supplementing the DeepSeek-R1-Llama8B results presented in the main text. 

Consistent with our primary findings, we observe that the latent knowledge signal is established early in the thinking trace and remains comparatively flat. Furthermore, the verbalization signal rises monotonically with depth, consistently overtaking knowledge partway through the sequence and peaking in the output trace.

\section{Steering}
\label{sec:appendix_steering}

We report additional steering results for Qwen3-30B-Instruct, Gemma4-31B, and LLaMA-3.1-8B, complementing the Qwen3-4B-Instruct results in the main text (\S\ref{subsec:rq5}).

\begin{figure*}[t]
    \centering
    \begin{subfigure}[b]{0.48\textwidth}
        \centering
        \includegraphics[width=\textwidth]{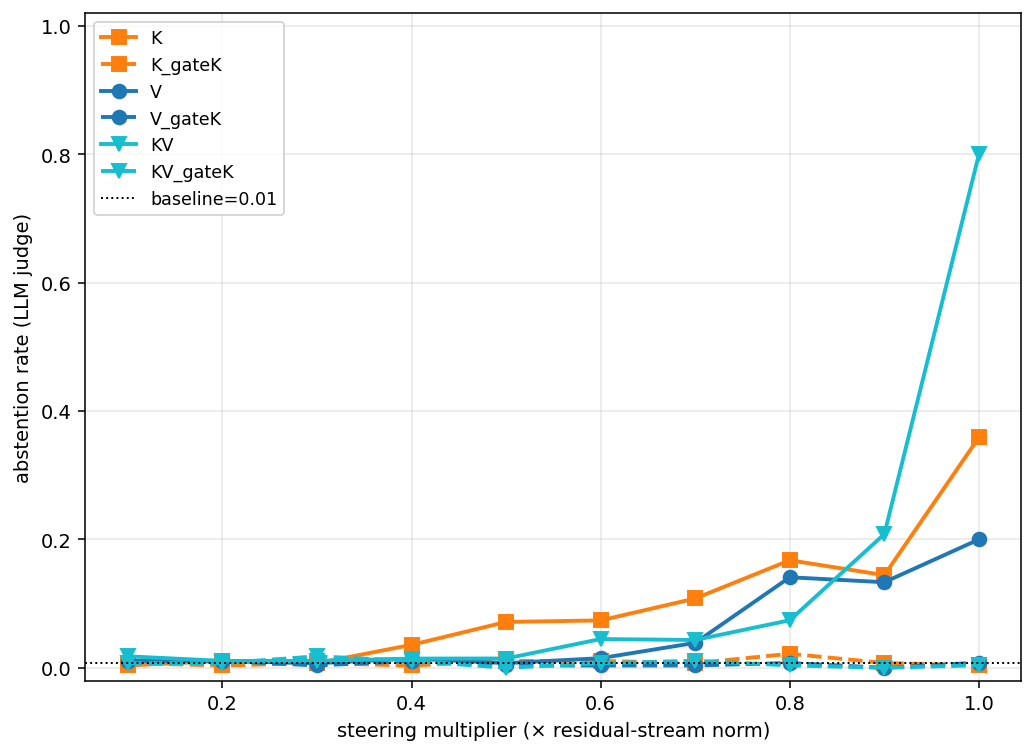}
        \caption{Solvable split (specificity control); lower abstention is better.}
        \label{fig:steer_solvable_30b}
    \end{subfigure}
    \hfill
    \begin{subfigure}[b]{0.48\textwidth}
        \centering
        \includegraphics[width=\textwidth]{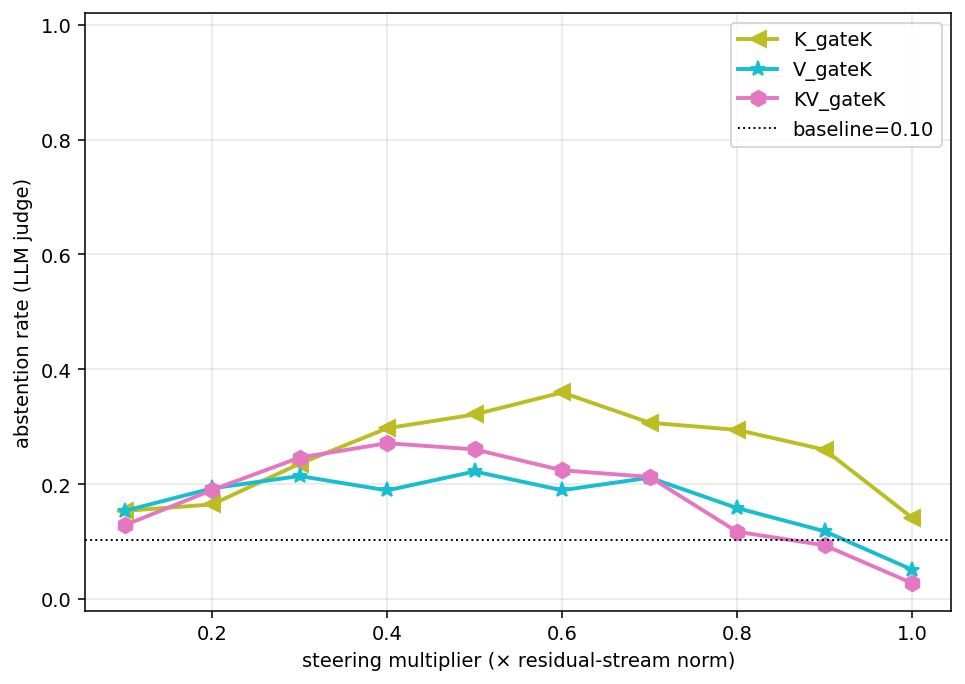}
        \caption{Unsolvable split; higher abstention is better.}
        \label{fig:steer_unsolvable_30b}
    \end{subfigure}
    \caption{Steering dose-response for Qwen3-30b-instruct, layer~36.}
    \label{fig:steer_30b}
\end{figure*}

\paragraph{Qwen3-30B-Instruct} At layer~36, results mirror the 4B model: ungated steering severely damages specificity on the solvable split (Figure~\ref{fig:steer_solvable_30b}), while gating restores low abstention across all three directions. On the unsolvable split (Figure~\ref{fig:steer_unsolvable_30b}), gated variants again rescue abstention, but with a different ranking: KgateK is now the strongest and most stable, outperforming KVgateK and VgateK at high multipliers, suggesting that at larger scale, gated steering on K alone suffices to rescue abstention on unsolvable problems.

\paragraph{Gemma and Llama} We attempted the same protocol on Gemma and Llama. Although probing confirmed the intervention shifted internal representations along the targeted K, V, and KV directions, these shifts had no causal effect on output: abstention rates stayed statistically indistinguishable from baseline across all multipliers, gated or ungated. This suggests the K/V decomposition, while representationally present, is not causally load-bearing for abstention in these models.

\begin{table*}[t]
\centering
\small
\begin{tabular}{lccc}
\toprule
\textbf{Model / Trace} & \textbf{V-only} & \textbf{Joint} & \textbf{K-only} \\
\midrule
Qwen3-4B-Instruct                  & 31.9 & 13.2 & 9.4 \\
Qwen3-30B-Instruct                 & 29.9 & 17.9 & 7.5 \\
Llama-3.1-8B-Instruct              & 35.9 & 20.9 & 14.9 \\
Gemma-4-31B-it                     & 57.4 & 8.3  & 7.5 \\
Qwen3-4B-T (response)              & 47.4 & 25.7 & 5.0 \\
Qwen3-4B-T (thinking)              & 28.6 & 12.2 & 11.0 \\
Qwen3-30B-T (response)             & 45.7 & 20.1 & 9.2 \\
Qwen3-30B-T (thinking)             & 32.2 & 15.0 & 11.6 \\
DeepSeek-R1-Llama8B (response)     & 44.2 & 12.5 & 8.0 \\
DeepSeek-R1-Llama8B (thinking)     & 19.5 & 7.7  & 12.8 \\
\bottomrule
\end{tabular}
\caption{Percentage of probe flips (Verbalization-only, Knowledge-only, Joint) under the conditioned prompt compared to the standard CoT baseline (unsolvable problems)}
\label{tab:conditioned_flip}
\end{table*}

\end{document}